\documentclass[10pt,twocolumn,letterpaper]{article}
 \usepackage{flushend}

\usepackage{cvpr}              
\definecolor{cvprblue}{rgb}{0.21,0.49,0.74}
\usepackage[pagebackref,breaklinks,colorlinks,allcolors=cvprblue]{hyperref}

\def\paperID{*****} 
\def\confName{CVPR}
\def\confYear{2026}

\title{Cross-Modal Corroboration for Annotation-Free Wildlife Monitoring}

\author{Bharath Pillai\\
The Ohio State University\\
{\tt\small pillai.105@buckeyemail.osu.edu}
\and
Varun Viswapriyan 
\and
Christopher Stewart\\
\and
Tanya Berger-Wolf
\and
Jenna Kline
}

\begin{document}
\maketitle
\begin{abstract}
Scaling wildlife monitoring for real-world conservation deployments requires automated analysis of smart sensors that operate under severe annotation scarcity.
We propose leveraging expert knowledge of species activity patterns as an annotation-free validation signal for multimodal monitoring pipelines.
We operationalize agreement as the alignment of independently derived hourly activity curves both with each other and with published behavioral priors~---~a three-way convergence that rules out shared-data confounds and dataset-internal correlation as alternative explanations.
Our vision pipeline combines zero-shot species detection via BioCLIP\,2, sliced inference to handle deployment-constrained camera positioning, and geometry-based geographic localization from camera trap imagery.
Our acoustic pipeline detects species vocalizations via a fine-tuned classifier.
We validate the pipeline on a breeding herd of Milu deer and demonstrate that both modalities independently recover activity patterns consistent with known deer behavioral ecology with minimal manual annotation.
The framework applies to species detectable in both visual and acoustic modalities for which behavioral priors are documented in the literature, suggesting a practical path toward self-validating wildlife-monitoring pipelines at conservation scale.
\end{abstract}

\section{Introduction}
\label{sec:intro}

\begin{figure}
    \centering
    \includegraphics[width=0.9\linewidth]{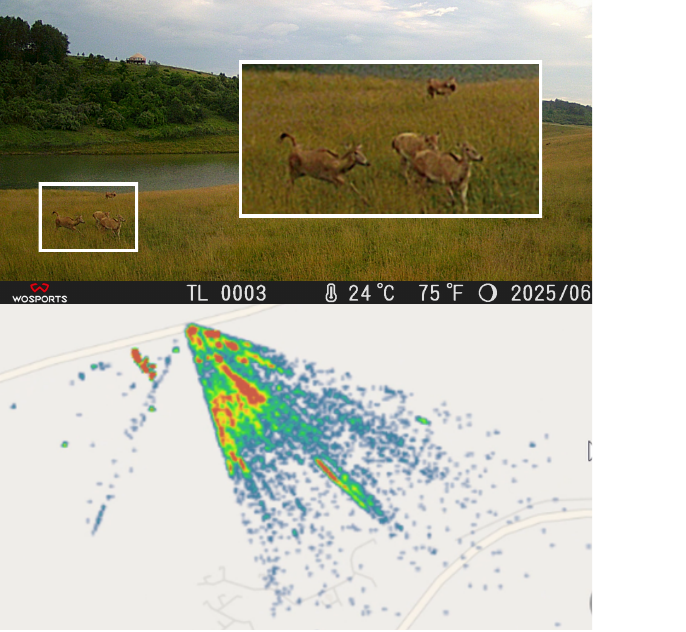}
    \caption{Top: Milu deer captured using a camera trap location 1 from the SmartWilds dataset \cite{Kline2025SmartWilds}. Bottom: Heat map of Milu deer detections from camera trap location 1.}
    \label{fig:hero}
\end{figure}

Wildlife monitoring at scale presents a dual challenge for computer vision.
Rare and endangered species yield few training examples by definition, while the cost of expert annotation makes labeling the data that does exist prohibitively expensive at conservation-relevant scales.
Camera trap networks and passive acoustic monitors can generate terabytes of data across vast geographic areas, yet expert
annotation remains a bottleneck that conservation programs rarely have resources to address~\cite{buxton2028pairing, tuia2022perspectives}.

However, domain experts do possess reliable prior knowledge about species-specific behavioral patterns: activity rhythms, habitat preferences, and the conditions under which vocalizations occur.
These behavioral priors, which are well-established in ethology even for data-scarce species, can serve as a lightweight supervision signal for validating deployed perception systems, requiring little to no manual annotation.
If two independently trained, modality-specific systems converge on the same activity pattern, and that pattern is consistent with known species ethology, this cross-modal agreement constitutes evidence that both systems are operating correctly.

Why does cross-modal agreement function as reliability evidence rather than mere correlation?
The argument rests on three independence properties.
First, the vision and acoustic pipelines share no model weights, training data, or input modality, so a coordinated failure would require two independent systems to produce the same incorrect signal.
Second, by comparing the converged signal against ethological priors documented in unrelated prior literature~\cite{cheng2025behavioral}, we eliminate the possibility that the convergence reflects a dataset-internal artifact: agreement between the two pipelines alone could be explained by shared environmental cues, but agreement that \emph{also} matches an externally documented diurnal pattern cannot.
Third, partial agreement~---~patterns appearing in one modality but not the other~---~is itself diagnostic, indicating behavioral states that one channel can capture but the other cannot, as we illustrate in Section~\ref{sec:discussion}.
Mere correlation between the two pipelines, without the external prior comparison, would not support the reliability claim.

We validate this framework as a multimodal monitoring pipeline validated on Milu deer (former name P\`ere David's deer, \textit{Elaphurus davidianus}), a critically endangered species, once extinct in the wild, with well-documented behavioral ecology~\cite{jiang2016elaphurus, cheng2025behavioral}.
We use the camera trap and acoustic recordings from the SmartWilds dataset~\cite{Kline2025SmartWilds}, monitoring a breeding group of Milu deer at The Wilds safari park \cite{thewilds2026}.
Our pipeline pairs a zero-shot vision module, BioCLIP 2~\cite{Gu2025BioClip}, for species detection in camera-trap imagery with a fine-tuned acoustic classifier for vocalization detection.
Rather than fusing modalities to improve detection accuracy, we use cross-modal corroboration, the agreement between the two independently derived activity signals, as an annotation-free reliability metric.

\section{Related Works}
Computer vision models for processing camera trap images have advanced rapidly, from general-purpose animal detectors such as MegaDetector~\cite{Beery_Morris_Yang_2019} to robust tracking and segmentation of video~\cite{carion2025sam, wasmuht2025sa}, to automated depth estimation~\cite{johanns2022automated}.
BioCLIP 2, trained on 214 million biological images, exhibits emergent properties in its embedding space that align with ecological and functional meaning, enabling strong zero-shot species recognition without per-species fine-tuning~\cite{Gu2025BioClip}.
For small or distant animals in wide-angle trap imagery, Sliced Aided Hyper Inference (SAHI) provides sliced inference that substantially improves the detection of low-resolution subjects~\cite{akyon_slicing_2022}.

Despite the complementary strengths of camera traps and acoustic recorders, few studies have combined them~\cite{buxton2028pairing}.
Growcott et al.\ \cite{growcott2024leopard} demonstrate that pairing passive acoustic monitoring with camera traps can enable individual identification of leopards via vocalizations, revealing ecological insights inaccessible to either modality alone.
MammAlps \cite{gabeff2025mammalps} similarly shows that multi-view video systems can support fine-grained behavioral monitoring of wild mammals, including deer.
In species distribution modeling, van Tiel et al.~\cite{tiel2024multi} demonstrate that late fusion of multimodal environmental data improves distributional predictions, though this work addresses plants rather than animal monitoring.

Our approach differs from prior multimodal wildlife work in a key respect.
Rather than using cross-modal data to improve a single inference task, we use cross-modal agreement as a proxy for system reliability.
This self-validating property is especially valuable when labeled evaluation sets are unavailable, which is often the case in conservation.

\section{Methodology}

\subsection{Dataset and Deployment}
Camera traps and acoustic recording units (ARUs) were deployed at The Wilds safari conservation Park (Cumberland, Ohio, USA) across six sites within a 220-acre pasture containing Milu deer~\cite{Kline2025SmartWilds}.
ARU and camera trap positions were recorded at deployment, along with camera trap bearing directions and horizontal field of view (HFOV). Image filenames encode acquisition timestamps in the format \texttt{YYMMDDHHMMSS}, which the pipeline parses automatically.
The dataset spans three sampling days between June 30 and July 7, 2025 (June 30, July 1, July 7), yielding 28{,}376 total camera trap detections and 5{,}792 acoustic recordings across all sites.
The sensor deployment locations were restricted to pre-existing structures, such as shade shelters and trees, to minimize disturbance to the habitat.

\subsection{Vision Pipeline}

The vision pipeline produces detection, classification, and geolocation data from camera-trap images in four stages.
Raw camera trap frames are processed through combined YOLO and SAHI detection.
BioCLIP\,2 is used to perform zero-shot classification.
Geometric distance and angle estimation, and GPS
forward projection produces per-detection spatial coordinates.

\paragraph{Stage 1: Object Detection (YOLO + SAHI).}
We use an Ultralytics YOLOv8 model as the base detector~\cite{yolov8}, integrated with the SAHI
framework~\cite{akyon_slicing_2022}.
SAHI divides each input image into overlapping
$512 \times 512$ pixel tiles (overlap ratio 0.2 in both dimensions), runs inference on each tile independently, and merges results using non-maximum suppression (IoU threshold 0.5).
Camera traps were mounted on existing
structures within the pasture, constraining positioning and frequently placing animals far from the camera. This sliced inference strategy substantially
improves detection recall for small or distant animals that would otherwise fall below the resolution threshold of standard full-image inference. Detections are filtered at a threshold of 0.25.

\paragraph{Stage 2: Zero-Shot Species Classification (BioCLIP 2).}
Each detected bounding box is cropped from the original image and passed to BioCLIP 2.
BioCLIP 2 is used in zero-shot classification mode via OpenCLIP, comparing image embeddings against text prompts of the form \textit{``a photo of a \textlangle label\textrangle''}.
For this deployment,
the label set was: \textit{white-tailed deer}, \textit{deer}, \textit{bird}.
The label with the highest cosine similarity score is assigned to each detection. Future deployments will include \textit{P\`ere David's deer} as an explicit label to improve taxonomic precision.

\begin{figure}
    \centering
    \includegraphics[width=\linewidth]{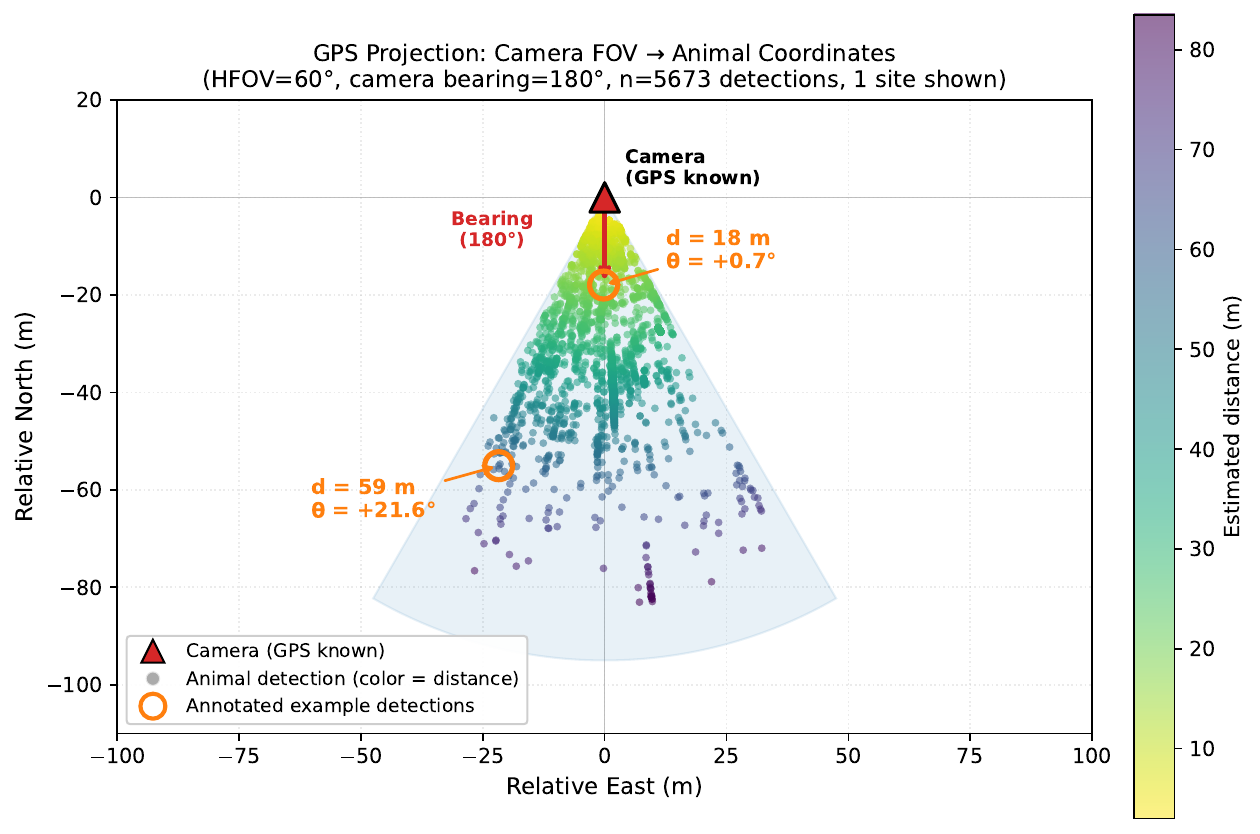}
    \caption{GPS projection model applied to real detections from one camera-trap site. The camera's known position (red triangle) and bearing ($180^{\circ}$, due south) define the field-of-view cone (shaded). Each scatter point is a single Milu deer detection, colored by estimated distance from the camera. Two annotated examples illustrate the bounding-box-to-coordinate mapping: a near-axis detection at $d=18$\,m, $\theta=+0.7^{\circ}$ and a wider-angle detection at $d=59$\,m, $\theta=+21.6^{\circ}$. The visible structure within the cone (e.g.\ depth-clustered points along radial lines) reflects spatial preferences of the herd within the camera's view rather than an artifact of the projection. \textit{Note:} This figure aggregates 5{,}673 detections from a single site; absolute distance accuracy is not validated against ground truth (see Section~\ref{sec:limitations}).}
    \label{fig2}
\end{figure}

\paragraph{Stage 3: Distance and Angle Estimation.}
Given a bounding box and known camera parameters, we estimate the distance and horizontal angle to each detected animal using a pinhole camera model, illustrated in Figure \ref{fig2}.
The focal length in pixels is derived from the camera's horizontal field of view (HFOV) and image width $W$:
\[
f_x = \frac{W/2}{\tan(\text{HFOV}/2)}
\]
Distance is estimated from the apparent height of the bounding box relative to the known real-world height of the target species ($h_{\text{real}}$, in meters):
\[
d = \frac{h_{\text{real}} \times f_x}{h_{\text{bbox}}}
\]
The horizontal angle relative to the camera center is computed as:
\[
\theta = \arctan\!\left(\frac{c_x - W/2}{f_x}\right)
\]
where $c_x$ is the horizontal center of the bounding box.
This approach assumes a flat ground plane and a known species height, and is most accurate for mid-range detections (10--150\,m).

\paragraph{Modeling assumptions and known limitations.}
\label{sec:limitations}
This pinhole formulation deliberately trades accuracy for simplicity and zero calibration overhead, and three assumptions break in real camera-trap deployments.
(i)~\emph{No lens distortion model.} Camera-trap optics typically exhibit modest barrel distortion that is most severe near the image edges; we treat this as a fixed-bias error term that grows with $|\theta|$ and recommend per-camera radial calibration for deployments where edge accuracy is critical.
(ii)~\emph{Bounding-box height as a proxy for animal height.} The bbox height equals the visible silhouette height, which is biased downward by occlusion (tall grass, vegetation) and upward by raised postures (rearing, alert stance).
For grazing species in pasture environments such as the Wilds site, mature deer in tall grass produce shorter bboxes than nominally expected, which the model interprets as greater distance~---~i.e., distance estimates in occluded conditions are biased \emph{long}.
(iii)~\emph{Flat-ground assumption.} The forward projection assumes the animal is at the same elevation as the camera; substantial terrain gradient between camera and subject introduces an additional bias proportional to $\tan(\Delta\text{elevation}/d)$.
We did not perform a quantitative calibration against ground-truth distance measurements; the absolute error of the geolocation module on this deployment is therefore unknown.
Per-detection coordinates should be interpreted as relative-position estimates suitable for downstream activity-space and home-range analyses rather than as absolute fixes.
Calibration against staged ground-truth targets is left to future work and would substantially improve confidence in the absolute output.

\paragraph{Stage 4: GPS Forward Projection.}
Given the camera's GPS coordinates $(\text{lat},\, \text{lon})$, recorded bearing, and the estimated distance and relative angle from Stage~3, we compute
the absolute bearing to the animal and project forward using the spherical Earth formula (WGS84):
\[
\text{absolute\_bearing} = (\text{camera\_bearing} + \theta) \bmod 360^{\circ}
\]
Projected animal coordinates are computed via Vincenty approximation, yielding per-detection \texttt{animal\_lat} and \texttt{animal\_lon} fields for direct integration with GIS tools. If the camera bearing is unavailable, the pipeline falls
back to the camera position as a conservative estimate.

\subsection{Acoustic Pipeline}

\paragraph{Stage 1: Positive Vocalization Mining.}
Training a reliable deer-call detector requires positive clips reflecting real field conditions.
Candidate positives are mined from unlabeled audio using
short reference segments containing confirmed Milu deer vocalizations, identified by domain experts.
All segments are standardized to 16\,kHz mono and
denoised with a low-pass filter to attenuate high-frequency artifacts (e.g., birds and insects).
Each segment is embedded using a pretrained Wav2Vec2
model~\cite{baevski2020wav2vec}; a mean-pooled embedding over final hidden states serves as a compact clip representation.
Reference embeddings are averaged into a prototype vector, and cosine similarity is computed against all candidate clips.
Clips exceeding a fixed similarity threshold are accepted as
additional positives after manual review, prioritizing precision over recall to reduce noise.

\paragraph{Stage 2: Feature Extraction and Classification.}
Each clip is represented using frame-level embeddings from
YAMNet~\cite{yamnet}, pooled over time by concatenating the mean and standard deviation across frames to capture both average spectral content and within-clip variability.
Feature vectors are standardized and reduced to 256 components via PCA before training.
A gradient boosted binary classifier~\cite{chen2016xgboost} is trained with class weighting to account for the imbalance between rare positive vocalizations and background clips.
Probability outputs are calibrated on a held-out split using sigmoid calibration, and the operating threshold is selected by maximizing F1 on the validation set.

\subsection{Cross-Modal Corroboration}

\paragraph{Activity signal construction.}
Both pipelines emit timestamped detection events.
A camera-trap ``detection'' is defined as a single bounding box surviving YOLO+SAHI non-maximum suppression at confidence $\geq 0.25$ \emph{and} receiving a BioCLIP\,2 zero-shot label of \emph{deer} or \emph{white-tailed deer}.
An acoustic ``detection'' is a 10-second clip whose calibrated positive-class probability exceeds the F1-optimal operating threshold (0.42 on the validation split, see Section~3.3).
Both event streams are aggregated into hourly bins; vision counts are summed across all six camera sites and across all deployment days, yielding a 24-element activity vector $\mathbf{v} = [v_1, \ldots, v_{24}]$.
The acoustic activity density $\mathbf{a} = [a_1, \ldots, a_{24}]$ is the per-hour fraction of clips at any ARU exceeding the threshold, similarly pooled across sites and days.
Figure~\ref{fig:activity} shows mean diurnal activity averaged over the deployment period; per-day camera-trap breakdowns are provided in supplementary Figure~\ref{fig:s1_per_day} and exhibit the same qualitative structure, indicating the pattern is not driven by a single anomalous day.

\paragraph{Reliability inference.}
High correlation between $\mathbf{v}$ and $\mathbf{a}$, in the absence of any shared model components or training data, constitutes annotation-free evidence that both systems are recovering the same underlying activity signal.
The recovered pattern is then compared qualitatively against the crepuscular and multi-peak diurnal rhythms documented for reintroduced Milu deer via GPS telemetry~\cite{cheng2025behavioral}; agreement with this external prior distinguishes genuine recovery from coincidental within-dataset correlation.

\section{Results}

\begin{figure*}
    \centering
    \includegraphics[width=0.85\linewidth]{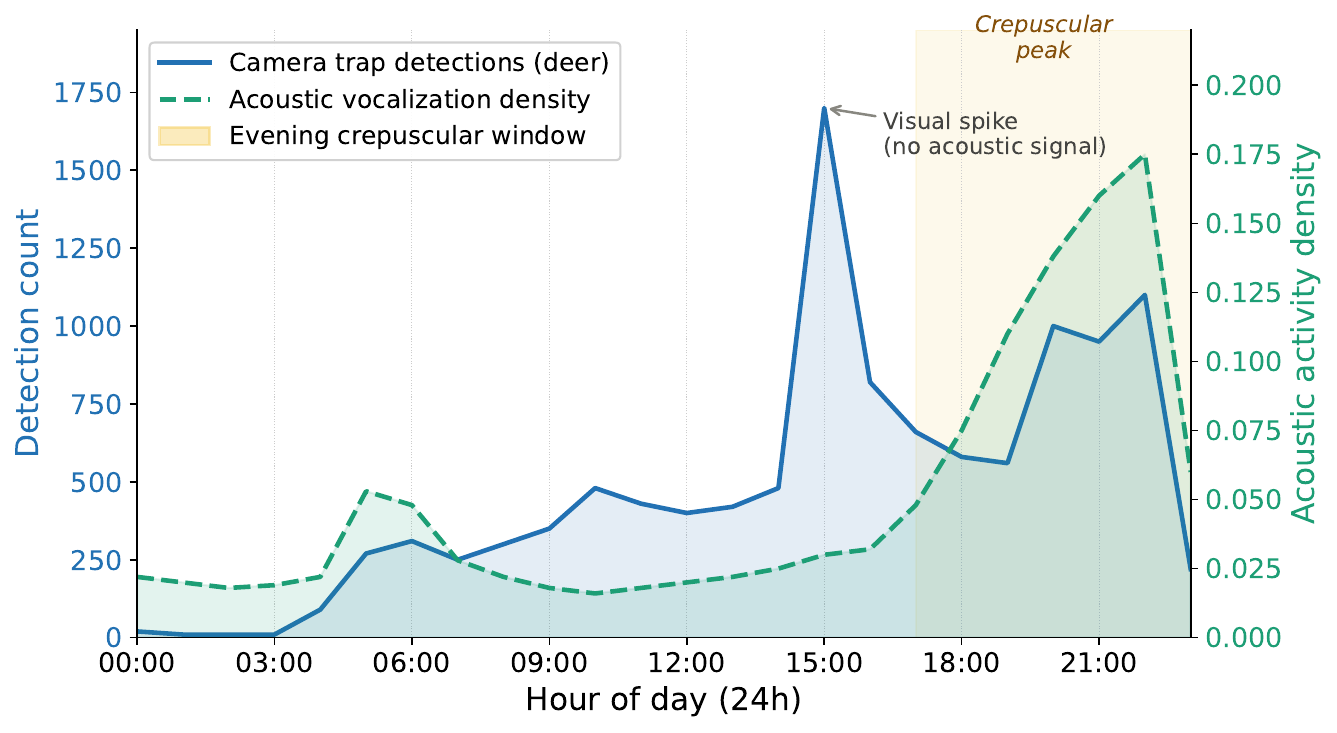}
    \caption{Mean diurnal activity pattern of Milu deer over the deployment, captured by camera traps and acoustic recordings. Camera-trap detections are defined per Section~3.4; acoustic activity density is the per-hour fraction of 10\,s clips above the calibrated F1-optimal threshold of 0.42. Both signals are pooled across all six sites and all deployment days. Annotated peaks~---~the mid-afternoon spike and the evening crepuscular peak~---~are discussed in Sections~4 and~5. Per-day camera-trap curves appear in supplementary Figure~\ref{fig:s1_per_day}.}
    \label{fig:activity}
\end{figure*}

The vision pipeline processed all images from the deployment, producing 28{,}376 animal detections.
Species breakdown by BioCLIP 2 zero-shot classification yielded 14{,}761 bird detections, 8{,}236 deer, and 5{,}379 white-tailed deer.
The two deer labels together account for 48\% of detections, consistent with the known composition of the pasture.
Mean YOLO detection confidence was 0.46 (SD 0.18), and mean BioCLIP 2 classification score for deer detections was 0.59 (SD 0.05).
Estimated distances ranged from 3\,m to 491\,m (mean$=$70\,m), broadly consistent with the spatial extent of the 220-acre pasture and known grazing locations.
Without ground-truth measurements, we cannot report absolute error, but two patterns in the distribution warrant interpretation.
First, the long tail of high-distance estimates concentrates at the periphery of the camera FOV, where bounding boxes are small, and the inverse-bbox-height distance estimator becomes increasingly noise-sensitive.
Second, the lower bound (3\,m) corresponds to large bounding boxes near the lens; these are the most reliable estimates and align with the range over which the pinhole approximation is well-conditioned.
We recommend treating per-detection distances over 150\,m as ordinal rather than metric for downstream spatial analyses.

The acoustic classifier achieves a positive-class precision of 0.963,
recall of 0.985, and F1 of 0.974 on the validation set, correctly
identifying 129 of 131 deer vocalizations while producing only 5 false
positives across 988 background clips. Combined with the semi-supervised
positive mining procedure, these results demonstrate that a small set of
expert-annotated reference calls are sufficient to train a high-precision
detector under severe class imbalance.

Hourly activity patterns (Figure~\ref{fig:activity}) reveal a pronounced
evening peak in deer detections (19:00--23:00) consistent with the crepuscular behavioral rhythms documented for reintroduced Milu deer via GPS telemetry
across multiple seasons~\cite{cheng2025behavioral}.
This convergence between an independently trained vision and acoustic pipelines and known ethological priors provides annotation-free evidence that the system is recovering genuine animal activity patterns rather than detection artifacts.
The acoustic pipeline independently recovers the same evening peak in vocalization density, further corroborating the signal across modalities.
A secondary spike in camera trap detections at approximately 15:00 is
absent from the acoustic signal, suggesting the two modalities are
capturing distinct behavioral states rather than a single shared activity
pattern.

\section{Discussion}
\label{sec:discussion}

Monitoring behavioral adaptation is critical for reintroduction programs,
where population recovery depends on successful integration into historical
habitat.
For Milu deer, understanding whether reintroduced individuals recover their natural activity rhythms is a core conservation question~\cite{cheng2025behavioral}, yet one that is difficult to answer at scale without animal-borne sensors.
Our pipeline produces spatially explicit, temporally resolved activity data
from passive sensors alone, enabling home range estimation and activity
space characterization without GPS collaring.

The 15:00 divergence illustrates the distinct role of the acoustic modality in this framework.
The acoustic signal is not introduced as a coverage mechanism; the visual pipeline already captures both the mid-afternoon and evening peaks.
Rather, the acoustic signal serves as the independent corroborating channel required for the reliability argument: it is the second leg of a three-way convergence with the published ethological prior.
The 15:00 spike is informative \emph{because} it appears in vision but not in audio.
Under our framework, this asymmetry is not a missed detection but a behavioral signature: mid-afternoon movement without active vocalization is consistent with the multi-peak activity structure documented for reintroduced Milu~\cite{cheng2025behavioral}, and such partial agreement, interpreted against the prior, distinguishes silent-locomotion states from vocalization-rich states.
A single-modality pipeline could not make this distinction; corroborated agreement on the crepuscular peak combined with informative disagreement on the afternoon peak is what supplies the reliability evidence.

This use of paired visual and acoustic sensors contrasts with prior multimodal monitoring work that fuses overlapping modalities to enrich a single behavioral inference.
MammAlps~\cite{gabeff2025mammalps}, for example, pairs multi-view video with audio to support fine-grained behavioral classification of wild mammals, leveraging overlap between modalities to refine the per-event label.
Our framework instead exploits the \emph{non-overlap} between visual and acoustic channels: agreement supplies reliability evidence, while disagreement is read as a behavioral signature against the ethological prior.
The two approaches are complementary~---~MammAlps demonstrates that overlapping modalities can be combined for richer behavioral characterization at the event level, while our framework demonstrates that the same modality pairing, when held \emph{independent}, can serve as an annotation-free validation signal at the population level.

\paragraph{Scope of the framework.}
The cross-modal corroboration approach has two practical preconditions.
First, the target species must be detectable in both modalities: a species that vocalizes rarely or whose vocalizations are masked by environmental noise will produce a degenerate acoustic curve regardless of pipeline correctness.
A primarily nocturnal, silent species would supply no acoustic signal to corroborate against.
Second, behavioral priors must exist in the published literature; while many ethological surveys exist for charismatic and reintroduction-focused species like Milu deer, data-scarce species often lack documented activity rhythms — indeed, generating such priors is frequently the goal of a camera-trap survey rather than its assumed input.
The framework, therefore, complements rather than replaces hypothesis-generating surveys: it is best suited to validating monitoring pipelines for species whose activity patterns are only partially understood, with each successful corroboration providing additional support for the prior literature.
\paragraph{Failure modes worth flagging.}
The independence-of-systems argument breaks under correlated failure, which we identify as the most serious threat to the reliability claim.
Two scenarios warrant explicit acknowledgment.
First, if two co-occurring species are misclassified by both pipelines in correlated ways~---~for example, a sympatric ungulate consistently labeled \emph{deer} by BioCLIP\,2 and producing acoustically similar vocalizations classified by the binary detector as Milu calls~---~the resulting activity curves can converge while neither system is operating correctly.
The single-species composition of the deployment pasture mitigates this risk in the present work but does not generalize.
Second, environmental confounds shared by both modalities~---~for instance, a diurnal anthropogenic disturbance that suppresses both visible and audible animal activity~---~could produce a false-negative pattern that is nonetheless convergent across modalities.
Mitigating these failure modes likely requires either species-level acoustic classification (rather than binary deer/non-deer) or a third independent signal source; both are promising directions for future work.

Several limitations point to future directions.
BioCLIP 2 performs
reasonably in zero-shot mode despite Milu deer being absent from the label
set; including the species' common and scientific names as explicit labels
is expected to improve precision and recall.
The label ambiguity between  \textit{deer} and \textit{white-tailed deer} could be resolved by  consolidating the label set.
A trade-off worth flagging without claiming to have measured it: sliced inference recovers smaller bounding boxes than full-image inference, which qualitatively suggests it improves detection recall on distant subjects (the design intent of SAHI) while also feeding the distance estimator inputs that lie further down the noise-amplification curve described in Section~3.2.
We did not run a controlled with/without-SAHI ablation in this work, so the magnitude of either effect is undetermined; quantifying the recall gain and any associated distance bias is a clear next step.
Calibration against ground-truth measurements would similarly improve absolute localization accuracy.
Finally, both pipeline modules are inference-only at test time,
making them natural candidates for edge deployment in the field~\cite{yu2024edge, vuilliomenet2026future}, enabling responsive, self-validating monitoring at scales impractical with cloud-dependent pipelines.

\section{Conclusion}
\label{sec:conclusion}

We presented a cross-modal corroboration framework for annotation-free validation of wildlife monitoring pipelines, demonstrated on Milu deer
using the SmartWilds dataset. Agreement between independently trained
vision and acoustic systems, grounded in known behavioral priors,
sidesteps the annotation bottleneck that limits scalable conservation
monitoring. Both pipeline modules are inference-only at test time,
making Edge AI deployment a natural next step toward
self-validating monitoring systems deployable at conservation scale.

\section{Data Availability Statement}
The data used in this study is available on \href{https://huggingface.co/collections/imageomics/smartwilds}{Hugging Face} \cite{Kline2025SmartWilds}.
The scripts are available on GitHub: \href{https://github.com/Imageomics/milu-deer-bioacoustics}{milu-deer-bioacoustics} \cite{viswapriyan2026wilds} and \href{https://github.com/Imageomics/camera-trap-geolocation}{camera-trap-geolocation}.

\section{Acknowledgments}

We thank Dan Beetem, The Wilds, and the Columbus Zoo \& Aquarium for their support in facilitating this project. All data collection was conducted under the supervision of the Director of Animal Management, with permission from The Wilds Animal Care and Use Committee. \\
\\
This project is supported by the \href{https://icicle.osu.edu/}{AI Institute for Intelligent Cyberinfrastructure with Computational Learning in the Environment (ICICLE)}, the \href{https://imageomics.org/}{Imageomics Institute}, and the \href{http://abcresearchcenter.org/}{AI and Biodiversity Change (ABC) Global Center}. ICICLE is funded by the US National Science Foundation under \href{https://www.nsf.gov/awardsearch/showAward?AWD_ID=2112606}{Award No. 2112606}, the Imageomics Institute is funded by the US National Science Foundation's Harnessing the Data Revolution (HDR) program under \href{https://www.nsf.gov/awardsearch/showAward?AWD_ID=2118240}{Award No. 2118240} (Imageomics: A New Frontier of Biological Information Powered by Knowledge-Guided Machine Learning). The ABC Global Center is funded by the US National Science Foundation under \href{https://www.nsf.gov/awardsearch/showAward?AWD_ID=2330423&HistoricalAwards=false}{Award No. 2330423} and the Natural Sciences and Engineering Research Council of Canada under \href{https://www.nserc-crsng.gc.ca/ase-oro/Details-Detailles_eng.asp?id=782440}{Award No. 585136}. This work draws on research supported by the Social Sciences and Humanities Research Council.

\section*{Sup plemental Material}

\begin{figure*}[h]
    \centering
    \includegraphics[width=0.95\linewidth]{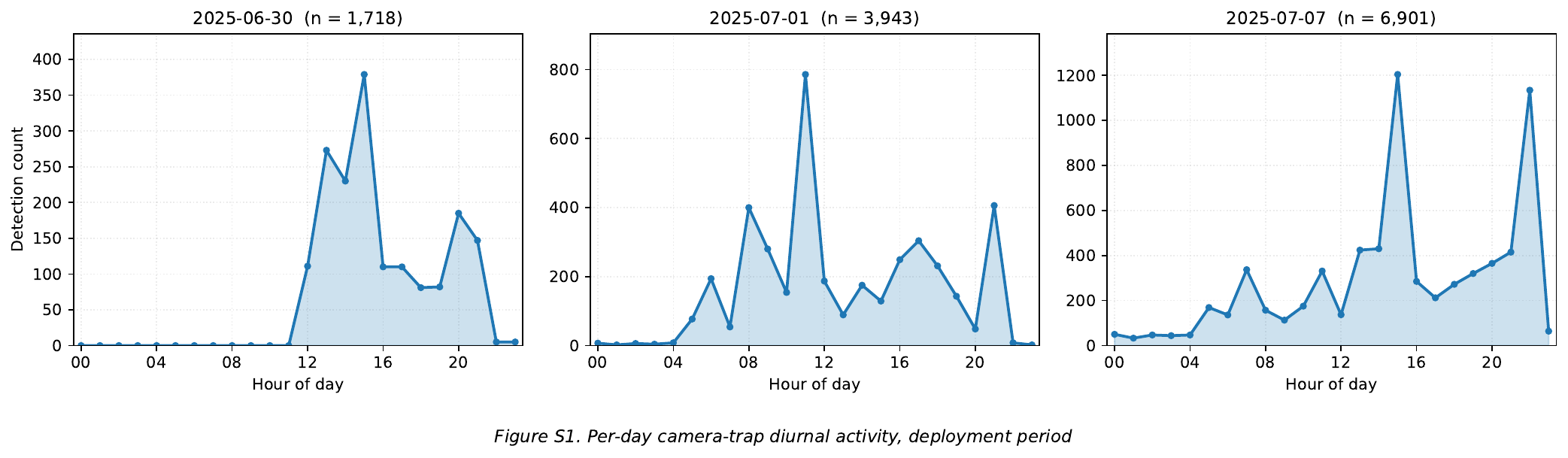}
    \caption{Per-day camera-trap diurnal activity over the deployment period. Each panel shows hourly detection counts for a single sampling day, with per-day detection totals noted in each panel title. The same qualitative structure~---~minimal nocturnal activity and elevated activity through midday into the late afternoon and evening~---~appears across all three days, confirming that the patterns shown in Figure~\ref{fig:activity} are not driven by a single anomalous day.}
    \label{fig:s1_per_day}
\end{figure*}

{
    \small
    \bibliographystyle{ieeenat_fullname}
    \bibliography{main}
}


\end{document}